\documentclass[sigconf,screen,nonacm]{acmart}
\newcommand{\tir}{TensorIR}
\newcommand{\block}{block}
\newcommand{\blocks}{blocks}
\newcommand{\tensorintrin}{TensorIntrin}

\usepackage[normalem]{ulem}
\usepackage{algorithm}
\usepackage{algorithmicx}
\usepackage{algpseudocode}
\usepackage{amsmath}
\usepackage{xcolor}
\usepackage{soul}

\begin{document}

\title{\tir: An Abstraction for Automatic Tensorized Program Optimization}


\author{Siyuan Feng}
\authornote{Both authors contributed equally to the paper}
\affiliation{
  \institution{Shanghai Jiao Tong University}
  \city{Shanghai}
  \country{China}
}
\email{hzfengsy@sjtu.edu.cn}

\author{Bohan Hou}
\authornotemark[1]
\affiliation{
  \institution{Carnegie Mellon University}
  \city{Pittsburgh}
  \country{USA}
}
\email{bohanhou@cs.cmu.edu}

\author{Hongyi Jin}
\authornote{Part of this work was done at Shanghai Jiao Tong University}
\affiliation{
  \institution{Carnegie Mellon University}
  \city{Pittsburgh}
  \country{USA}
}
\email{hongyij@cs.cmu.edu}

\author{Wuwei Lin}
\affiliation{
  \institution{OctoML}
  \city{Seattle}
  \country{USA}
}
\email{wlin@octoml.ai}

\author{Junru Shao}
\affiliation{
  \institution{OctoML}
  \city{Seattle}
  \country{USA}
}
\email{jshao@octoml.ai}

\author{Ruihang Lai}
\authornotemark[2]
\affiliation{
  \institution{Carnegie Mellon University}
  \city{Pittsburgh}
  \country{USA}
}
\email{ruihangl@cs.cmu.edu}

\author{Zihao Ye}
\affiliation{
  \institution{University of Washington}
  \city{Seattle}
  \country{USA}
}
\email{zhye@cs.washington.edu}

\author{Lianmin Zheng}
\affiliation{
  \institution{UC Berkeley}
  \city{Berkeley}
  \country{USA}
}
\email{lmzheng@berkeley.edu}

\author{Cody Hao Yu}
\affiliation{
  \institution{Amazon Web Services}
  \city{Seattle}
  \country{USA}
}
\email{hyuz@amazon.com}

\author{Yong Yu}
\affiliation{
  \institution{Shanghai Jiao Tong University}
  \city{Shanghai}
  \country{China}
}
\email{yyu@apex.sjtu.edu.cn}

\author{Tianqi Chen}
\affiliation{
  \institution{Carnegie Mellon University, OctoML}
  \city{Pittsburgh}
  \country{USA}
}
\email{tqchen@cmu.edu}
\email{tqchen@octoml.ai}

\renewcommand{\shortauthors}{Siyuan and Bohan, et al.}

\newcommand{\etal}{\textit{et al}.}
\newcommand{\ie}{\textit{i}.\textit{e}.}
\newcommand{\eg}{\textit{e}.\textit{g}.}

\newcommand{\precap}{\vskip -0mm}
\newcommand{\postcap}{\vskip -0mm}
\newcommand{\presec}{}
\newcommand{\postsec}{}

\newcommand{\Note}[1]{}
\DeclareRobustCommand{\NoteSigned}[3]{{\sethlcolor{#2}\Note{#1: #3}}}
\newcommand{\NoteSiyuan}[1]{\NoteSigned{siyuan}{pink}{#1}}
\newcommand{\NoteBohan}[1]{\NoteSigned{bohan}{yellow}{#1}}
\newcommand{\NoteJunru}[1]{\NoteSigned{junru}{yellow}{#1}}
\newcommand{\NoteTianqi}[1]{\NoteSigned{TQ}{green}{#1}}
\newcommand{\NoteLianmin}[1]{\NoteSigned{lianmin}{green}{#1}}
\newcommand{\NoteCody}[1]{\NoteSigned{cody}{green}{#1}}

\newcommand{\algorithmautorefname}{Algorithm}

\begin{abstract}

Deploying deep learning models on various devices has become an important topic. 
The wave of hardware specialization brings a diverse set of
acceleration primitives for multi-dimensional tensor computations.
These new acceleration primitives, along with the emerging machine learning models, bring tremendous engineering challenges.
In this paper, we present \tir{}, a compiler abstraction for optimizing
programs with these tensor computation primitives.
\tir{} generalizes the loop nest representation used in existing machine learning compilers
to bring tensor computation as the first-class citizen.
Finally, we build an end-to-end framework on top of our abstraction to automatically 
optimize deep learning models for given tensor computation primitives. Experimental results show
that \tir{} compilation automatically uses the tensor computation primitives for given hardware backends and delivers performance that is competitive to state-of-art hand-optimized systems
across platforms.

\end{abstract}





\maketitle
\section{Introduction}

\begin{figure}[t]
    \centering
    \includegraphics[width=\linewidth]{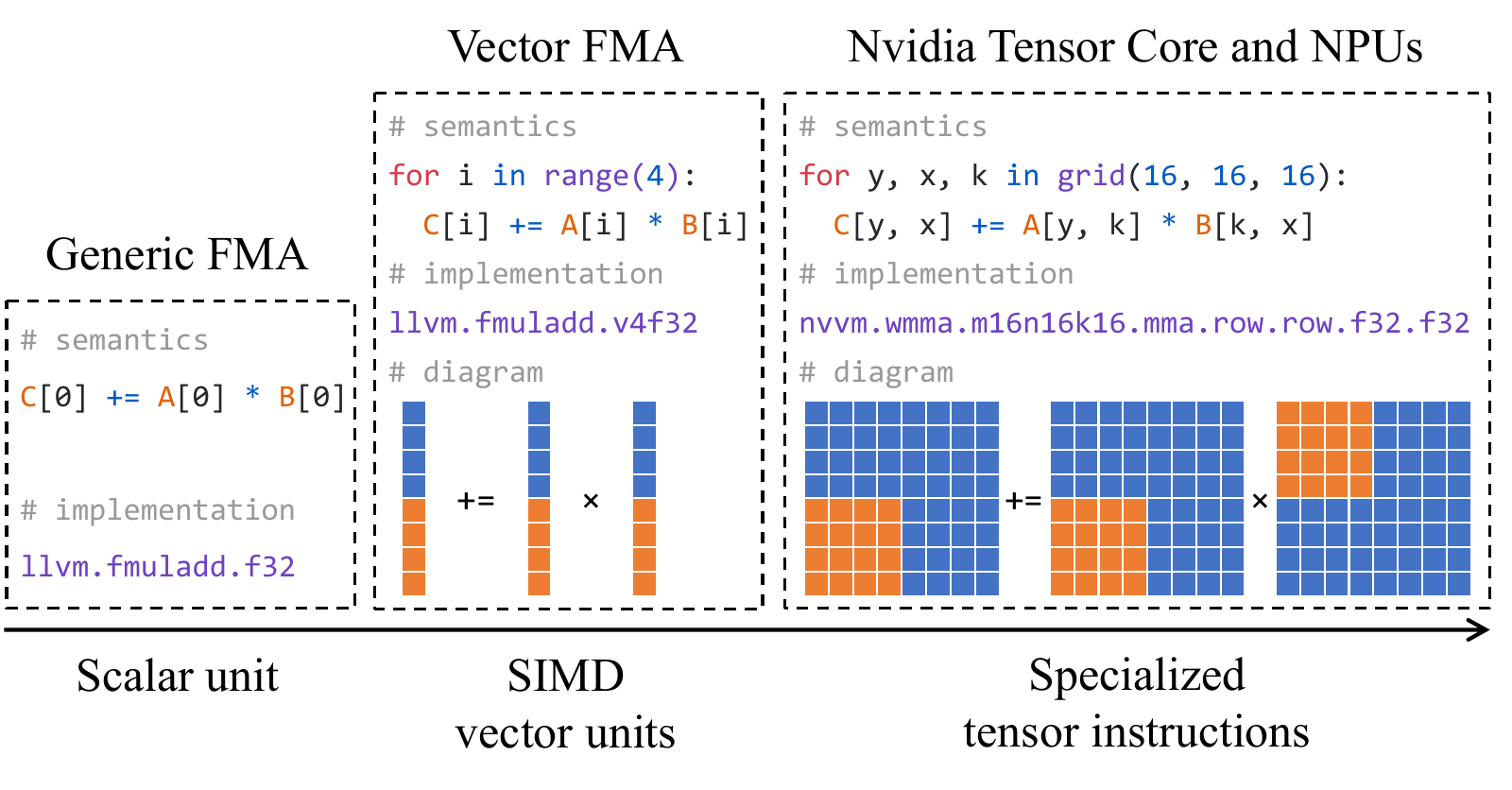}
    \precap
    \caption{Trends of hardware specialization. The classical acceleration technique uses vector units to process multiple scalar computations simultaneously, which is still widely used on CPU platforms. However, to cater to increasingly heavier computation throughput requirements, modern accelerators usually contain specialized high-dimensional tensor computation instructions, creating the need for tensorized program optimization.} 
    \postcap
    \label{fig:hw-trend}
\end{figure}

Deploying high-performance machine learning models has become an emerging challenge in various areas, including image recognition \cite{he2016deep, simonyan2014very, howard2017mobilenets}, natural language processing \cite{vaswani2017attention, devlin2018bert, radford2019language}, and games \cite{mnih2013playing, lillicrap2015continuous, schulman2017proximal}. The advances in machine learning bring demands to support a broad range of models. In the meantime, there are increasing demands to deploy smart applications to a broad spectrum of devices ranging from servers to embedded environments.
 
The wave of hardware specialization further complicates the problem~(\autoref{fig:hw-trend}). Driven by the goal of machine learning acceleration, modern hardware backends introduce specialized primitives to speed up tensor computations (e.g., Nvidia Tensor Core~\cite{nvidia2017tensorcore}, Google TPU~\cite{jouppi2017datacenter}). 
Domain experts also start to develop micro-kernel primitives, which carefully organize a series of highly optimized instructions to perform a sub-computation to speed up domain-specific tensor operator libraries.
These hardware instructions and micro-kernel primitives typically operate on multi-dimensional tensor regions and effectively perform tensor operations such as multi-dimensional loads, dot product, and matrix multiplication~(\autoref{fig:hw-trend}).
We call these opaque tensor computation acceleration constructs \emph{tensorized intrinsics} and transformation procedure to use these intrinsic \emph{tensorization}. 
In order to get the best out of these hardware backends, modern machine learning systems need to optimize programs that contain hierarchical loop nests, multi-dimensional loads, and tensor intrinsics -- we call this problem \emph{tensorized program optimization}.

Most of the current tensorized programs are optimized by domain experts, who compose tensorized primitives together with multi-dimensional loops, threading patterns, and data caching to craft specialized kernel libraries such as Intel MKL-DNN~\cite{intel2017mkldnn}, ARM Compute Library~\cite{acl} and NVIDIA cuDNN~\cite{chetlur2014cudnn}. 
These libraries are then used by machine learning frameworks such as TensorFlow~\cite{abadi2016tensorflow}, PyTorch~\cite{paszke2019pytorch} and MXNet ~\cite{chen2015mxnet}. However, huge engineering efforts are required to support the growing sets of models and backends, and it takes iteration cycles for these libraries to adapt to the rapidly changing and growing machine learning applications, which hinders the evolution of new machine learning models.

\begin{figure*}[!t]
    \centering
    \includegraphics[width=0.85\linewidth]{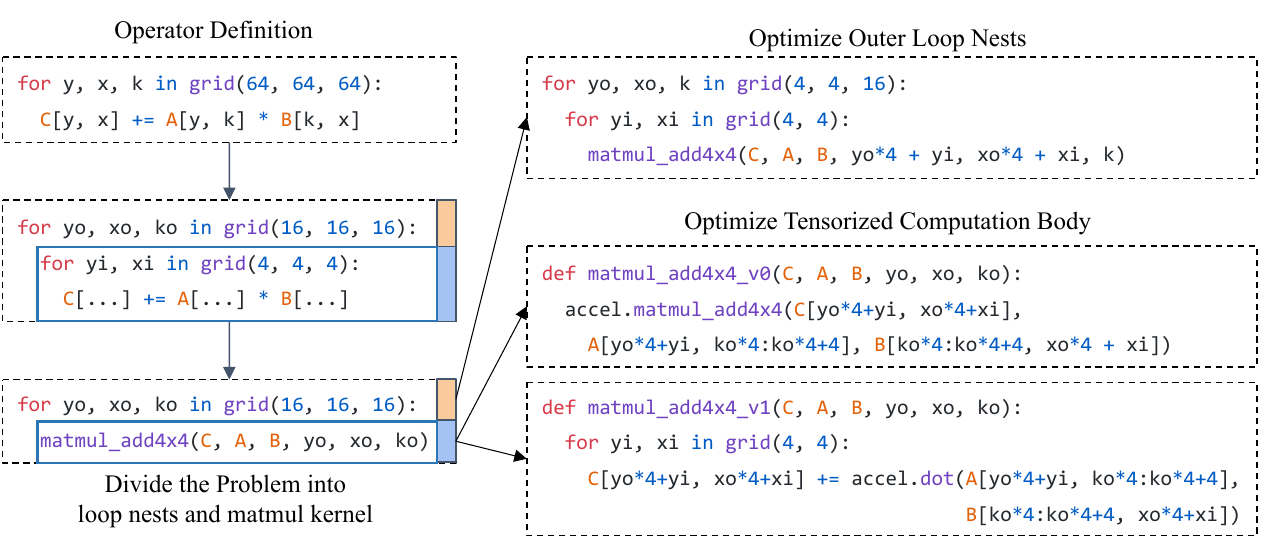}
    \precap
    \caption{An expert developer can choose to divide the problem into 4x4 matmul and loops that uses the 4x4 matmul, then optimize them separately. This way we can effectively make use of specialized tensor instructions in the target hardware.}
    \postcap
    \label{fig:motivation}
\end{figure*}

In this paper, we propose to address the tensorized program 
optimization problem using a automatic compilation approach. 
Most past works in machine learning compilation~\cite{chen2018tvm, vasilache2018tensor} search over a program space of loop nest transformations and do not handle tensorized programs automatically. Bringing automatic program optimization to tensorized programs would unlock the benefits from domain-specific accelerations in modern hardware backends.  We identify the following key challenges to achieving this goal: 
 

\paragraph{Abstraction for Tensorized Programs} 
To build an automated compiler for tensorized programs, we need an abstraction
that can pragmatically capture possible equivalent tensorized computations for a given machine learning operator. Notably, the abstraction needs to represent multi-dimensional memory accesses, threading hierarchies, and tensorized computation primitives from different hardware backends. The abstraction also needs to be expressive enough to represent most of the operators of interest in machine learning. 

\paragraph{Large Design Space of Possible Tensorized Program Optimizations} 
Another challenge is to produce an optimized tensorized program for a given operator automatically. A compiler needs to make use of a rich set of techniques that domain experts might use, including making effective use of loop tiling, threading, and data layout transformations. Importantly, these transformations now need to be made in conjunction with tensorized computations, bringing additional complexities to analysis and automation. The combinations of these transformations form a large search space. We need an effective way to find an optimized tensorized program for a given search space.

To address these challenges, we introduce \emph{\tir{}}, an abstraction for automatic tensor program optimization. To begin with, we
introduce a new construct called \emph{block} that allows us to divide and isolate 
tensorized computation region from the outer loop nests.
The new abstraction allows us to effectively represent tensorized computations and combine them with loop nests, threading, and memory hierarchy.
We also introduce program transformation primitives to express a 
rich space of potential optimizations.
We build a novel automatic scheduling algorithm 
on top of the abstraction and transformation primitives.
Additionally, \tir{} abstraction also allows us to represent and optimize
programs that contain a mixture of irregular computations
and tensor computations, expanding the possible support beyond a normal tensor expression~\cite{chen2018tvm}. This paper makes the following contributions: 


\begin{itemize}
    \item We propose a novel abstraction for tensorized programs that separates tensorized computation from the loop transformations. Meanwhile, the same abstraction allows us to uniformly represent tensor intrinsics and hardware constraints.
    \item We build transformation primitives to generate a rich search space of tensorized program optimization with correctness validation.
    \item We design and implement a new tensorization-aware automatic scheduler.
\end{itemize}
We integrate \tir{} with an end-to-end compilation framework and show that it outperforms existing machine learning compilation solutions by up to 7x and automatically brings competitive performance to heavily optimized platform-specific solutions.


\section{Overview}
\label{sec:overview}

\begin{figure*}[htb]
    \centering
    \includegraphics[width=\linewidth]{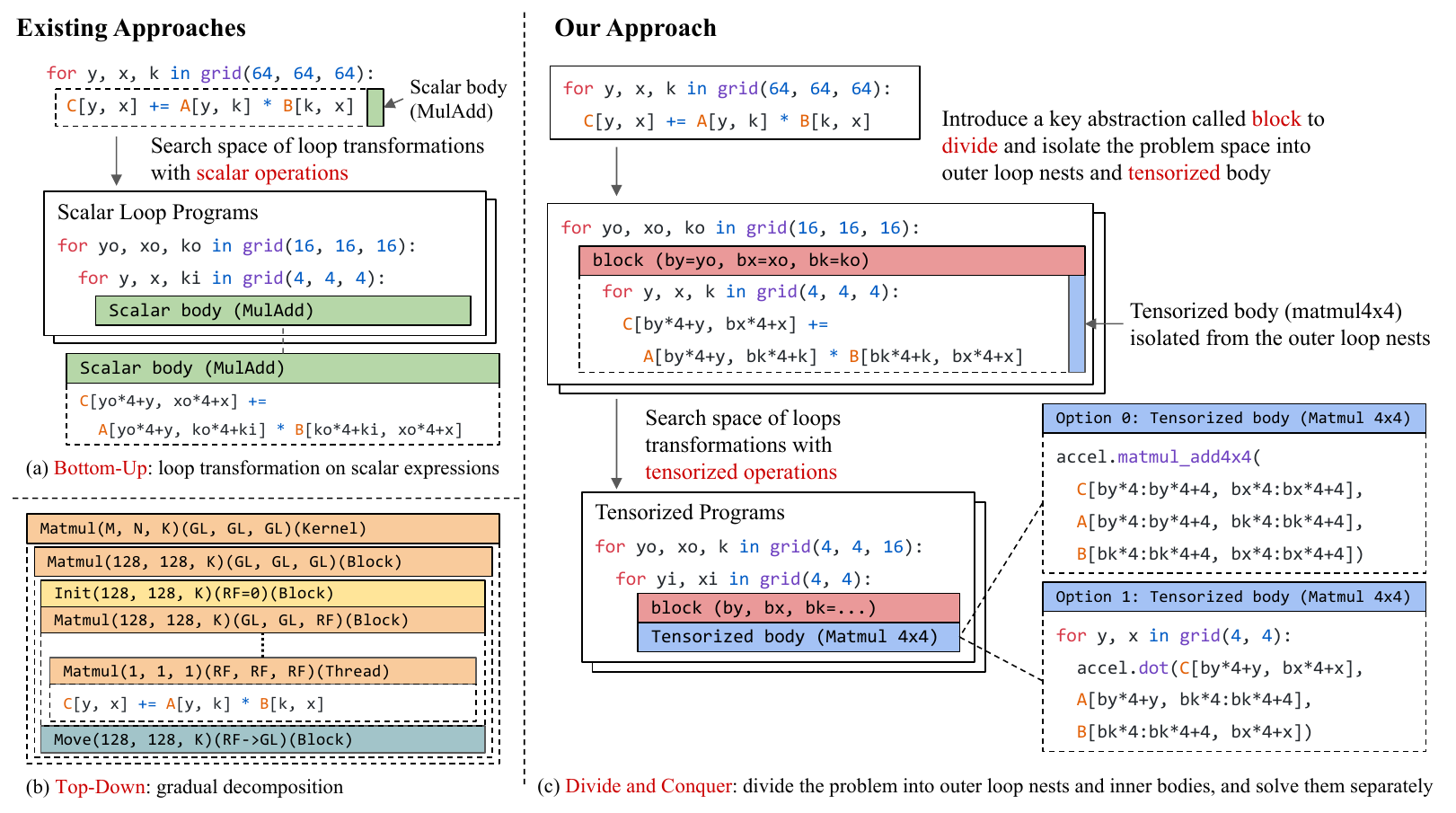}
    \precap
    \caption{Overview of our approach. We use a key abstraction named block to divide and isolate the tensorized computations, and enables further loop transformations with tensorized operations.}

    \postcap
    \label{fig:overview}
\end{figure*}

This section describes the key insights of our approach and gives an overview of the paper.
To motivate our approach, we start with an example flow of how a domain expert optimizes a tensorized program in \autoref{fig:motivation}. Tensorized computation primitives usually correspond to a sub-problem of the original tensor operator computation.
As a result, it is natural for domain experts to choose a \emph{divide and conquer} approach
-- divide the original program into sub-problems of tensorized computation and loop nests
that use the tensorized computation, then optimize them separately.
The divide and conquer approach allows developers to focus on a sub-problem without worrying
about the others. Additionally, it also enables us to target multiple tensorized computation implementations.

Most existing machine learning compilers take two kinds of approaches~(\autoref{fig:overview}).
Halide~\cite{ragan2013halide}, TVM~\cite{chen2018tvm}, Tiramisu~\cite{baghdadi19tiramisu}, AKG~\cite{zhao2021akg}, MLIR/Affine~\cite{mlir} and AMOS~\cite{zheng2022amos} take
a bottom-up approach that models the search space using loop nests iterators around scalar operation bodies, and then optimizes the program by finding the best loop nest transformation~(through search or polyhedral optimization). 
HTA~\cite{HTA}, Fireiron~\cite{hagedorn2020fireiron} and Stripe~\cite{zerrell2019stripe} use a top-down approach
that gradually decomposes the problem into sub-problems through nested polyhedral structures.
Given the significance of the divide and conquer approach in manual tensorized program optimizations, it is natural to ask whether it is possible to bring the same insight to machine learning compiler design.

We give a positive answer in this paper. Specifically, we introduce a new abstraction called \emph{\block{}} into the loop nests. 
A \block{} contains the right amount of signature information to \emph{isolate} the inner problem space and outer problem.
With \block{}, we can continue to perform transformations on both outer and inner problem independently, using the \block{} signature as the interface.
Similar to the manual divide and conquer approach, a common use case of a \block{} is to represent a tensorized computation primitive in a hardware backend, but we can also use the \block{} to isolate bigger sub-problems of interest when divide and conquer makes sense.
Importantly, tensor computation is the first-class citizen in \tir{}. 
Loop nests with \blocks{} can be viewed as a generalized abstraction of iteration space.
We present the detailed design of the \tir{} abstraction in \autoref{sec:tir}.

To automate the tensorized program optimization, we construct a search space of possible ways to divide the problem guided by the hardware tensor computation primitives, then further search over possible ways to solve sub-problems using program transformations. 
We present the automatic scheduling algorithm for tensorized programs in \autoref{sec:auto}.

Our divide and conquer covers the search space of previous compiler approaches (bottom-up, top-down) and generalizes the typical optimization techniques in HPC and ML engineering to an abstraction that allows automatic tensorization. We automate decisions common in library and compilation pipeline, enabling us to automatically generate competitive solutions with vendor-specific libraries. We present the experiment results in \autoref{sec:evaluation}.

\section{\tir{} Abstraction} \label{sec:tir}


\begin{figure}[t]
    \centering
    \includegraphics[width=\linewidth]{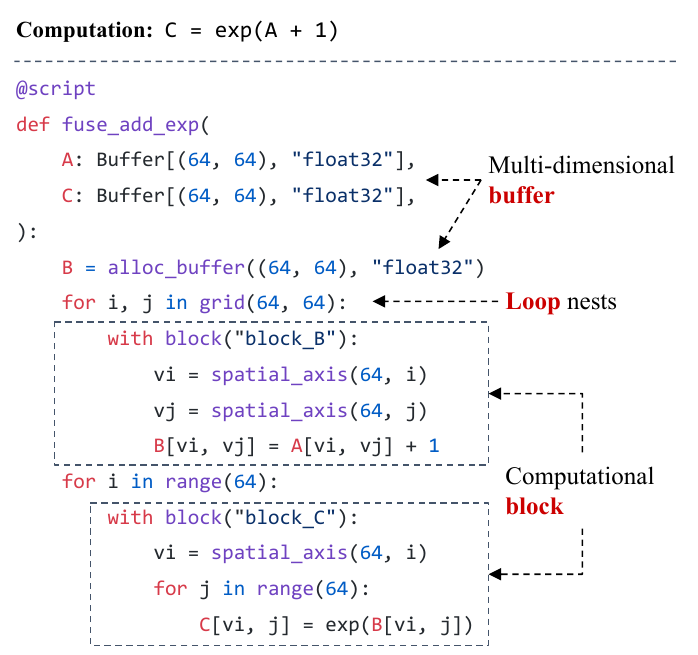}
    \precap
    \caption{An example \tir{} program with three major elements - multi-dimensional buffers, loop nests and computational \block{}. Details of \block{} is omitted for simplification.}
    \postcap
    \label{fig:elemnts}
\end{figure}

This section introduces the \tir{} abstraction. \autoref{fig:elemnts} gives an example of \tir{} program. We introduce a Python-AST (abstract syntax tree) dialect of \tir{} to let developers directly construct and transform programs in Python.
A \tir{} program contains three main elements: multi-dimensional buffers, loop nests~(with possible thread bindings in GPU settings), and \blocks{}. A \block{} can contain one or more nested loop nests with sub-\blocks{}
or sequence of imperative statements that correspond to the content of the computation. This representation allows us to divide computations into the corresponding sub(\block{})-regions and do effective program transformations
using dependency information stored in the block signature. We discuss the design details in \S\ref{subsec: block}.



\begin{figure}[t]
    \centering
    \includegraphics[width=\linewidth]{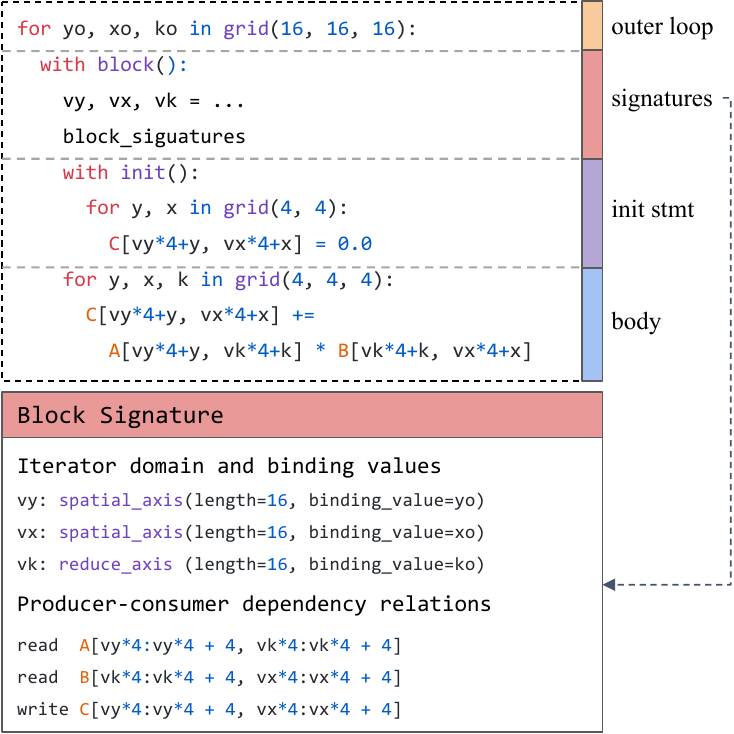}
    \precap
    \caption{Blocks contain complete signature for dependency analysis and we make it an isolation level between body computation and outer loop nesting.}
    \postcap
    \label{fig:block}
\end{figure}

\subsection{Block}
\label{subsec: block}

A block in \tir{} represents a tensorized computation on a sub-region of the multi-dimensional buffers.
\autoref{fig:block} shows an example block for the matrix multiplication (matmul) computation. 
The body of a block is parameterized by a set of block iterator variables $v_y, v_x, v_k$ which represents an abstract tensorized computation. Instantiated with different value combinations of these block iterator variables, the block maps to different concrete running block instances. These iterator variables can be bound to expressions that contain the outer loop iterators, which implies the execution order of block instances.

\paragraph{Rationale}
The main design rationale of a block is to isolate tensorized computation -- we 
want to be able to transform loop nests outside the block without looking into its body.
However, unlike scalar computation, we may not be able to extract the dependency information needed for transformation from an opaque tensor computation body.
As a result, we introduce a block signature that
contains sufficient dependency information for transformations.
We discuss these transformations in \S\ref{subsec:schedule}.
Additionally, the signature can be used to independently verify the correctness 
of the iterator bindings during transformations~(more details in \S\ref{subsec:validation}).

\paragraph{Block Iterator Domain}
While it is possible to instantiate a block's body computation by binding the block iterators to any loop nests, most instantiations do not correspond to the same computation.
To ensure the consistency of computation among transformations, 
we store the iterator domain information and the constraints
of iterators in the block signature. For the particular example
in \autoref{fig:block}, we know that $vx$, $vy$ and $vk$ must bind to iterators in domain $[0, 16)$. Additionally, because $vk$ is a reduction axis, we know that
we cannot bind it to a parallel loop unless the reduction is atomic.
The domain constraints still leave massive room for outer loop transformations,
as there are multiple ways to construct loops that satisfy the constraint. 
Our domain signature can be viewed as a specific way to represent the
integer domain sets and relations of the iterators. We choose the particular
representation due to its implementation efficiency and simplicity in reasoning,
but would also point out that the same design philosophy applies
to other formal domain representations of integer sets and relations~\cite{vasilache2006polyhedral}.

\paragraph{Access Region and Dependency}
To provide sufficient dependency information, a block signature contains the access regions
and read/write dependencies that a block has with respect to the multiple dimensional buffers.
In \autoref{fig:block}, the block writes the region $C[vy*4: vy*4 + 4, vx*4: vx*4 + 4]$ by reading $A[vy*4: vy*4 + 4, vk*4: vk*4 + 4], B[vk*4: vk*4 + 4, vx*4: vx*4 + 4]$.
The dependency information is used during transformations.
We only mark each block's dependency with respect to the multi-dimensional buffers instead of other statements~(blocks). This indirection enables a broader range of transformations, such as 
data layout transformation and re-computation which are essential in tensorized program optimization.


\paragraph{Reduction Block and Initialization}
A reduction computation usually contains an initialization step and an update step.
We can naturally map the reduction computation into two blocks. But, on the other hand, it is usually helpful to jointly make the scheduling decisions~(such as tiling and computation location) of the two steps. 
We introduce an optional initialization statement for blocks that perform reduction. 
An initialization statement is executed during the first iteration of a reduction.
This reduction block representation is mainly useful during transformations.
We provide transformation primitives to transform between the two-block-based representation
and the init-block-based representation so we can pick the best representation for low-level
code generation.

\begin{figure}[!t]
    \centering
    \includegraphics[width=0.75\linewidth]{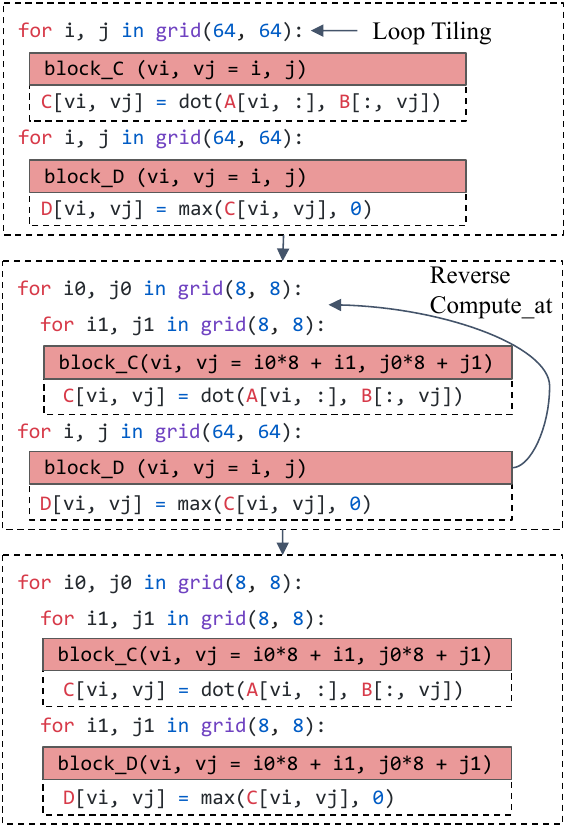}
    \precap
    \caption{Loop transformations mutate outside loop nests but change nothing inside the block.}
    \postcap
    \label{fig:schedule-loops}
\end{figure}

\begin{figure}[!t]
    \centering
    \includegraphics[width=0.75\linewidth]{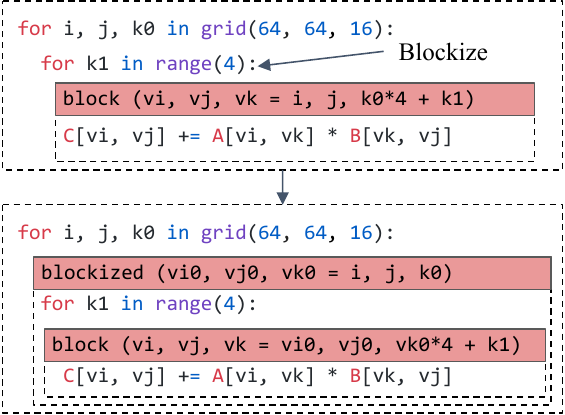}
    \precap
    \caption{Blockization creates a new block to isolate inside computation and outside loop nesting.}
    \postcap
    \label{fig:blockize}
\end{figure}

\subsection{Scheduling Transformations}
\label{subsec:schedule}

\begin{figure*}[t]
    \centering
    \includegraphics[width=\linewidth]{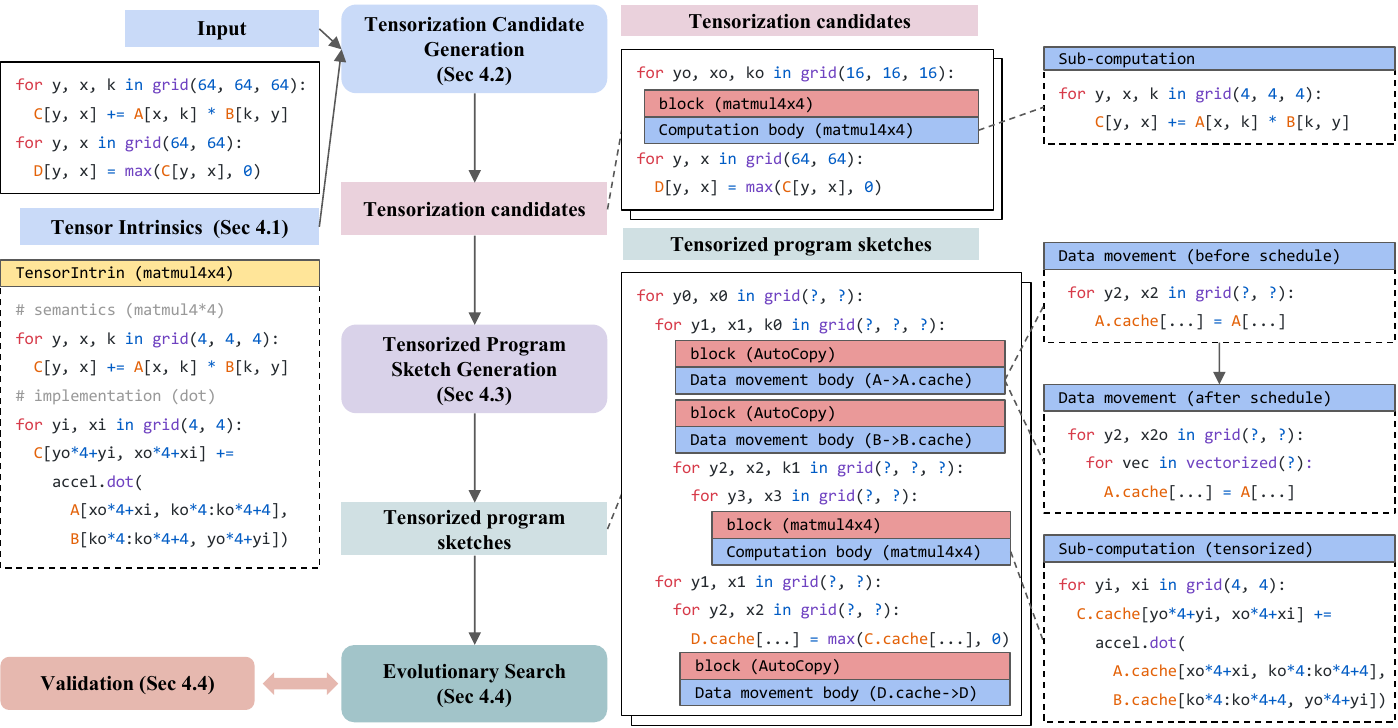}
    \precap
    \caption{Automatic optimization for tensorized program with hardware intrinsics. We take 64x64x64 matrix multiplication followed by a RELU operator as the input workload and 4x4x4 matmul as the synthetic tensor intrinsic which is implemented by a dot product instruction. The tensorization candidate generation step tiles the 64x64x64 GEMM into 4x4x4 sub-tiles and isolate the sub-computation. Then the tensorized program sketch generation step schedules the computation and insert the resulting data movement (AutoCopy) blocks which are scheduled independently. Finally, we use evolutionary search to fill the random decisions in sketches with a validation mechanism to filter out incorrect programs.
    }
    \postcap
    \label{fig:auto-tensorize}
\end{figure*}

For a given input program, we need to generate a rich search space
of programs with equivalent semantics. We introduce primitives
to transform a \tir{} program to equivalent optimized programs. Following the existing convention
of tensor program optimizations~\cite{ragan2013halide,chen2018tvm,baghdadi19tiramisu}, we call this procedure scheduling.

A block is schedulable if it only contains loop nests with sub-blocks as its leaves.
We can transform the loop nests and sub-block computation locations within a schedulable block
by analyzing the sub-block signatures and their dependency information.
Notably, a schedulable block can contain non-schedulable sub-blocks~(e.g., opaque Tensor Core computation). An opaque block can also contain a schedulable sub-block. 
Based on the block isolation, we can still effectively explore the search space of the schedulable part independently while keeping the same opaque block. 
We describe the schedule primitives in the rest part of this subsection.




\paragraph{Loop Transformations}

Loop transformations such as loop tiling ~(split, reorder) and compute location mutation are important ways to optimize programs for better memory locality.
We also provide these loop transformation primitives~(see examples in \autoref{fig:schedule-loops}).
Unlike existing tensor compilers that directly extract the dependency of each leaf
scalar computation statement, we calculate the dependencies by only inspecting the block signature.
Besides loop transformations, we also support primitives that bind
loops to GPU threads and provide annotation hints such as vectorization and unrolling. Note that the block isolation does not prevent many important collaborative optimization across blocks (e.g. inlining, cooperative fetching). Our loop transformations cover the loop transformations provided by previous works which allows \tir{} to reproduce their search space as mentioned in \autoref{sec:overview}. 

\paragraph{Blockization}

The loop transformation primitives preserve the overall hierarchy of 
blocks. As we alluded in \autoref{sec:overview}, sometimes dividing the problem by isolating a sub-region computation into a new sub-block is helpful. 
We call this transformation \emph{blockization} \autoref{fig:blockize}.
A blockized program is no longer scalar-based as the new sub-block corresponds
to a tensorized computation. We can use blockization to isolate possible candidates
for tensorization.
Besides blockization, we also introduce primitives that can change the block hierarchies. For example, we provide caching primitives that introduce sub-blocks to cache
input data into shared memory. We also provide back and forth transformations between a single reduction block and the corresponding init- and update blocks.

\paragraph{Separation of Scheduling and \tir{}}
Many previous tensor compilers~\cite{ragan2013halide, chen2018tvm} rely on a declarative 
scheduling language to construct a schedule tree. 
Adding new scheduling primitives to these compilers requires changes to both
the schedule tree data structure and the corresponding lowering rule in these compilers.
We take a different approach and implement each schedule primitive as a standalone transformation from one \tir{} program 
to another. This design is easier to extend,
as different developers can develop new primitives concurrently based on a stable \tir{} abstraction.  Additionally, developers can print out the program at any  
transformation stage for debugging and mix automatic rewriting with schedule transformations.




\subsection{Validation}
\label{subsec:validation}
The blocks and their buffer read/write relations capture a complete picture of the original computation and are used to validate the correctness of loop nests and threading assignments. 

\paragraph{Loop Nest Validation}
Loop nest validation checks whether the iterator binding provided by the loop nests matches the constraints of the iterator domain, including the domain size and iterator independence information. For example, if two data-parallel block iterators are bound as $v_1 = i; v_2 = i * 2$, then the corresponding program is invalid because $v_1$ and $v_2$ are not independent. 
But $v_1 = i / 4; v_2 = i \% 4$ can be a legal binding. We build 
pattern-matchers to find a quasi-affine mapping from the loop iterators to the block 
iterator variables and use the pattern to validate the independence and domain of the bindings.
Besides the iterator domain validation, it is also important to check the producer-consumer relations to make sure producer blocks that write to buffer regions always cover the read region of downstream consumers.

\paragraph{Threading Validation}
When building a program for GPUs and other accelerators with threading support, we also need to do additional validations
with respect to the threading and memory hierarchies. We do three kinds of validations: 
\begin{itemize}
    \item  \textbf{Thread binding}: Ensure different iterators bound to the same thread are consistent and meet the launching constraints of the backend. 
    \item \textbf{Cooperative memory access}: For blocks that produce buffers stored in shared memory collaboratively across threads, we need to ensure the block covers downstream requirements from all the threads in the same group. Meanwhile, upstream blocks that provide inputs for this block need to cover the read requirement of this block from all the threads in this group.
    \item \textbf{Execution scope}: Validate that tensor intrinsic runs at the correct execution scope~(e.g., TensorCore needs to run at the warp-level).
\end{itemize}


\paragraph{Correctness of Schedule Primitives}
We add checks to each schedule primitive to ensure the correctness of the transformation.
When a schedule primitive only changes the loop nests, we can also use the validation procedure to ensure correctness. Because the block iteration domains and dependencies stay the same in these cases. We find primitive-specific necessary conditions for schedule primitives that change the blocks~(e.g., blockization).




Note that loop nest validation and threading validation are used as checks to filter out invalid \tir{} programs and schedule primitive checks are used to ensure the equivalence of \tir{} programs before and after transformations. Users will get warning or error information when they are incorrectly manually crafting, importing and scheduling \tir{} programs. When users use the compiler to generate programs automatically which will be discussed in section \ref{sec:auto}, validation can help filter out false positive cases during the exploration in the search space. Hence, both user programs and compiled programs will benefit from the validation.

\subsection{Programming Effort} 
As shown in \autoref{fig:elemnts}, we provided a Python-AST dialect of \tir{} to allow developers directly construct, dump, inspect, modify, and transform \tir{} programs in Python. The program effort will be high if users need to specify all the computation and optimizations manually. Our framework allows users to import models from TensorFlow/PyTorch and automatically generates \tir{} programs from the high-level operators. Additionally, the system automatically provides the optimizations, such as tiling and caching, for a given hardware platform through the auto-scheduling (Section \autoref{sec:auto}). We still allow users to write \tir{} in Python dialect when they want customized operators. In these cases, the system provides optimization transformations automatically. As a result, the programming effort is usually minimized.

\begin{figure*}[t]
    \centering
    \includegraphics[width=\linewidth]{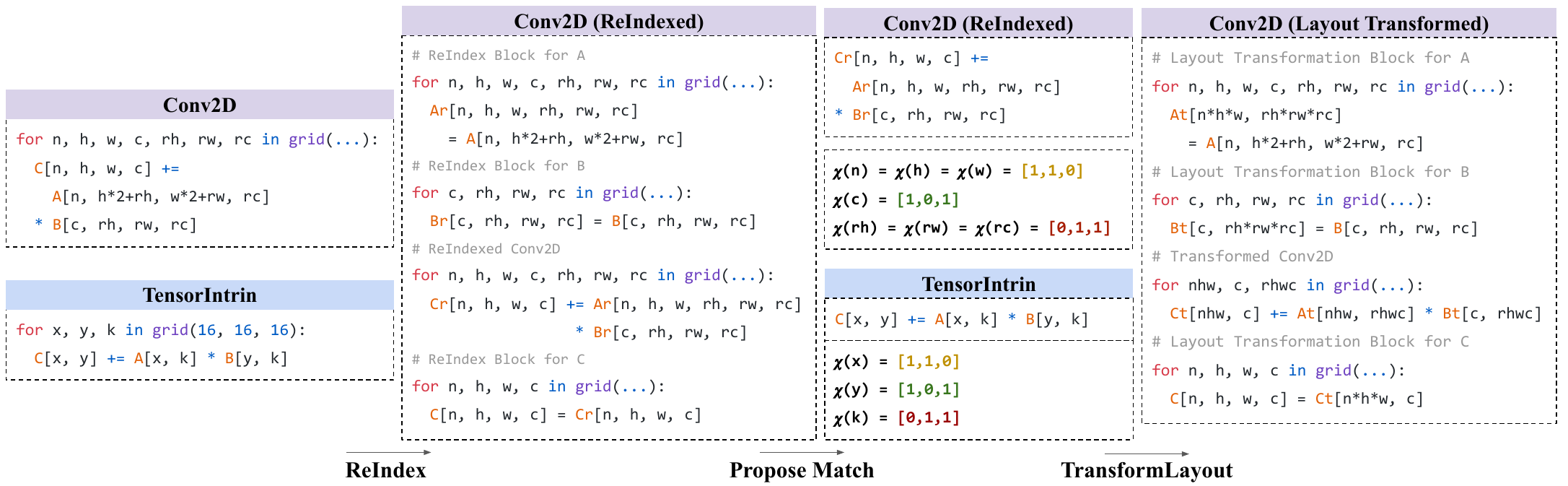}
    \precap
    \caption{Example flow of tensorization candidate generation. We take standard NHWC 2D convolution as the input workload and 16x16x16 matrix multiplication as the hardware backend intrinsic. First, the system converts the buffer access expressions to intermediate iterators. Based on the buffer access patterns, we calculate characteristic functions for each iterator and build a mapping between iterators that share the same characteristic vector. The mapping further guides the transformation of block instance space and \textit{ReIndex} buffers. Note that although the \textit{ReIndex} stages of B and C are redundant, they will be inlined into consumers during the sketch generation phase and as a result do not affect the performance.}
    \postcap
    \label{fig:matching}
\end{figure*}

\section{Auto-Scheduling Tensorized Programs} 
\label{sec:auto}


In the last section, we introduced \tir{} abstraction and a set of transformation primitives. 
In order to fully make use of the set of improvements, we need an automatic solution
to optimize over a set of transformations and map computations to 
the native tensor intrinsics. In this section, we describe a tensorization-aware automatic scheduler to solve this problem.

\autoref{fig:auto-tensorize} shows an overview of our approach. Our system takes a workload description from users and tensor intrinsic descriptions about the hardware platform as inputs. The auto-scheduler first generates candidates for tensorization by inspecting the computation pattern.
It then generates program sketch candidates that use the tensorized computations and then decide the data movements according to the compute patterns.
For a given search space induced by the tensorized program sketches, we perform evolutionary search guided by a learning-based cost model. 
The entire process centers itself around the tensorization and leverages the new block abstraction to isolate tensorized computations. 
We discuss the details of each step in the subsequent subsections.

\subsection{Abstraction for Tensor Intrinsics}

To make use of a tensor intrinsic in our optimization, we need a way to
provide its semantics and backend implementation to the system. 
We leverage the \emph{same} \tir{} abstraction to describe the tensor intrinsics of a given hardware backend.
For each tensorized instruction, we introduce a \tensorintrin{} construct composed of two blocks. One block describes the computation \textit{semantics}, and the other provides the low-level \textit{implementation} of the tensorized computation.

In \autoref{fig:auto-tensorize}'s example, we use a normal loop nest with scalar 
body $C[i, j] \mathrel{{+}{=}} A[i, k] * B[k, j]$ to represent the computation semantics
and implement the intrinsic using inner dot product instruction $accel.dot$.
We also include the data type, storage scope, memory layout, and contiguity
constraints through the multi-dimensional buffer specification in a \tensorintrin{}.
Those constraints are used during the validation step.

Notably, tensor intrinsics are usually applied together with special memory scopes, data layouts, and corresponding load/store instructions in common platforms. For instance, on NVIDIA GPUs, if we decide to use \textit{nvcuda::wmma::mma\_sync} API to perform dense computation, then we need to apply \textit{nvcuda::wmma::load\_matrix\_sync} and \textit{nvcuda::wmma::store\_matrix\_sync} to prepare input operands and retrieve output results respectively.~On ARM CPUs, micro-kernels like \textit{a64\_gemm\_u8\_8$\times$12} require operands to be stored in interleaved layout. Developers can inform the system about these constraints by specifying special storage scopes for each input and output operands of the tensor computation.


\subsection{Tensorization Candidate Generation}


Given a pair of backend target and an input program, we first match the program body
to possible \tensorintrin{} to generate tensorization candidates. 
 The match is performed in a gradual way. We first match the expression pattern $C[.] \mathrel{{+}{=}} A[.] \times B[.]$ without considering the indices. We then refine the matches by proposing possible mappings between the indices. 
\autoref{fig:matching} gives an example that walks through the matching process. In this example, we take a matrix multiplication intrinsic as the backend description.
The computation of this tensor intrinsic can be described by the following formula
\begin{equation}
    C[x,y] \mathrel{{+}{=}} A[x,k] \times B[k,y]
\end{equation}
It is easy for us to match the tensor intrinsic to workloads like batch matrix multiplication, which can be described by
$$ C[b,i,j] \mathrel{{+}{=}} A[b,i,r] \times B[b,r,j],
$$ 
by mapping $x, y, k$ to $i, j, r$. But for workloads like 2-dimensional Convolution~(Conv2D) with more complex index expression patterns
\begin{align*}
    C[n, h, w, co] \mathrel{+}= &A[n, h * s_h + r_h * d_h, w*s_w+r_w*d_w, r_c] \\ \times &B[r_c, r_h, r_w, co],
\end{align*}
the mapping between $x, y, k$ and $n, h, w, c, r_h, r_w, r_c$ is not straightforward. 

On these more general cases, we rewrite the computation expression into an equivalent form:
\begin{align*} 
    C[n, h, w, co] \mathrel{+}= A_r[n, h, w, r_h, r_w, r_c] \times B[r_c, r_h, r_w, co],\\
    A_r[n, h, w, r_h, r_w, r_c] = A[n, h * s_h + r_h * d_h, w*s_w+r_w*d_w, r_c].
\end{align*}
We call this transformation \textit{ReIndex} which uses intermediate iterators that appear in the buffer access indices to rewrite the buffer access expressions. To match the new computation to the tensor intrinsic, we check the buffer access where each iterator appears. For example, we notice that $n, h, w$ and $x$ appear in indices of $A(Ar), C$, $co$ and $y$ appear in indices of $B, C$, and $r_h, r_w, r_c$ and $k$ appear in indices of $A, C$. We can then match the iterators in the computation to the iterators in the tensor intrinsic by inspecting their appearance patterns. Specifically, we map $\mbox{fuse}(n, h, y)$ to $x$, $co$ to $y$, and $\mbox{fuse}(r_h, r_w, r_c)$ to $k$. Here $\mbox{fuse}()$ is to fuse multiple iterators together and can be recursively defined by
\begin{align*}
    \mbox{fuse}(i_1) &= i_1 \\
    \mbox{fuse}(i_1, i_2, \dots, i_{r}) &= \mbox{fuse}(i_1, i_2, \dots, i_{r-1}) * \mbox{extent}(i_{r}) + i_{r},    
\end{align*}
where $\mbox{extent()}$ is the extent of iterator $i_r$. We can then transform the computation to
\begin{align*}
    C_t[\mbox{fuse}(n, h, w), co] \mathrel{+}= &A_t[\mbox{fuse}(n, h, w), \mbox{fuse}(r_h, r_w, r_c)] \\ \times &B_t[\mbox{fuse}(r_h, r_w, r_c), co],
\end{align*}
where
\begin{align*}
    C_t[\mbox{fuse}(n, h, w), co] = &C[n, h, w, co], \\
    A_t[\mbox{fuse}(n, h, w), \mbox{fuse}(r_h, r_w, r_c)] = &A_r[n, h, w, r_h, r_w, r_c] \\
    B_t[\mbox{fuse}(r_h, r_w, r_c), co] = &B[r_h, r_w, r_c, co].
\end{align*}
We use this mapping to reshape the block instance space and the outer loops and transform the layout of \textit{ReIndex} buffers. We insert layout rewrite blocks to rewrite $A, B, C$ to $A_t, B_t, C_t$ respectively and use $A_t, B_t, C_t$ to rewrite the computation body. After these steps, the computation body is compatible with the tensor intrinsic. 


\paragraph{Loop Reorganization and Early Blockize}
Besides the computation body, we also need to ensure that the tensor computation region matches the description provided by the \tensorintrin{}. The shape of the reorganized block instance space might not be divisible by the sub-problem size of the tensor intrinsic. For each computation body from the last step, we do necessary padding on the computation block and input/output operands to the closest divisible shape. We then perform tiling to create inner loops to match the loop nest of the tensor intrinsic and further blockize the inner loop to isolate the corresponding tensor computations.
Notably, the candidates generated in this step do not always lead to successful tensorizations. This is because other constraints, such as memory layout
and threading depend on later transformations. These constraints are embedded in the tensorization candidates and checked during validation.

\paragraph{Formal Description of the Process} So far we gave a high-level overview of the tensorization candidate generation process. In the the remainder part of this subsection, we provide a formal description of the process. Suppose the intrinsic scalar expression can be formalized as
\begin{equation} \label{intrin}
    O[\textbf{v}_{0}] = f(O[\textbf{v}_{0}], I_1[\textbf{v}_1], I_2[\textbf{v}_2], \dots, I_k[\textbf{v}_k]).    
\end{equation}
where $O$ is the output operand, $I_{[1: k]}$ are input operands, $\textbf{v}$ is the set of iterators that parameterized this computation, $\textbf{v}_{[0:k]}$ are all lists of iterators that belong to $\textbf{v}$, and $f$ is the expression pattern detected. Note that it accommodates common dot product and matrix multiplication intrinsics. Furthermore, suppose that the workload scalar expression can be formalized as
\begin{align*}
    \tilde{O}[g_{0}(\tilde{\textbf{v}}_{0})] = f(\tilde{O}[g_{0}(\tilde{\textbf{v}}_{0})], \tilde{I}_1[g_1(\tilde{\textbf{v}}_1)], \dots, \tilde{I}_k[g_k(\tilde{\textbf{v}}_k)]),
\end{align*}
where $\tilde{O}$, $\tilde{I}_{[1: k]}$, $\tilde{\textbf{v}}_{[0:k]}$ corresponds to their counterparts and $f$ is exactly the same. $g_{[0:k]}$ are mappings that map lists of iterators to the actual buffer access position. In our Conv2D case, for instance, we have
$
    g_{A}(n, h, w, r_h, r_w, r_c)=(n, h * s_h + r_h * d_h, w*s_w+r_w*d_w, r_c).
$

To reduce the problem to a simpler canonical form, we apply the \textit{ReIndex} schedule transformation, which creates an intermediate cache buffer for an operand but with the layout changed according to the iterators. Formally, if we run \textit{ReIndex} $ \tilde{I}_1[g_1(\tilde{\textbf{v}}_1)]$, we create the following rewrite block before the computation
\begin{align*}
    \hat{I}_{1}[\tilde{\textbf{v}}_1] = \tilde{I}_1[g_1(\tilde{\textbf{v}}_1)].
\end{align*}
Then if we apply \textit{ReIndex} to all the operands, the workload scalar expression is reduced to
\begin{equation} \label{compute}
    \hat{O}[\tilde{\textbf{v}}_{0}] = f(\hat{O}[\tilde{\textbf{v}}_{0}], \hat{I}_1[\tilde{\textbf{v}}_1], \dots, \hat{I}_k[\tilde{\textbf{v}}_k]),
\end{equation}
where buffer access indices in both \ref{intrin} and \ref{compute} directly correspond to iterators.

To match \ref{intrin} and \ref{compute}, we define the characteristic vector $\chi(v) \in \{0,1\}^{k+1}$ of an iterator $v \in \textbf{v}$ by inspecting whether each of $\textbf{v}_0, \textbf{v}_1, \textbf{v}_2, \dots \textbf{v}_k$ contains $v$. Formally,
\begin{align*}
    \chi(v)_i &= [v \in \textbf{v}_i] \quad i \in [0, k], 
\end{align*}
where $[]$ is the Iverson bracket that returns 1 if the corresponding condition is true or 0 otherwise (\autoref{fig:matching}). We can successfully propose the mapping as long as $\forall v \in \textbf{v}, \exists \tilde{v} \in \tilde{\textbf{v}}, \chi(v) = \tilde{\chi}(\tilde{v})$. In the current implementation, we can further safely assume that iterators in $\textbf{v}$ all have different characteristic vectors. Then for all $v \in \textbf{v}$, we fuse all such $\tilde{v}$ where $\chi(v) = \tilde{\chi}(\tilde{v})$ and map the fused iterator to $v$.

Notably, the iterator order inside each of $\textbf{v}_{[0:k]}$ or $\tilde{\textbf{v}}_{[0:k]}$ does not affect the value of characteristic function $\chi$ or $\tilde{\chi}$. But when we fuse all $\tilde{v}$ where $\chi(v) = \tilde{\chi}(\tilde{v})$, the order of fusion affects how the operands are reorganized in memory to be compatible with the tensor intrinsic. Our implementation now uses a default order for all the workloads and can generalize to different fusion orders in the future.




\subsection{Tensorized Program Sketch Generation}
For a given set of tensorization candidates,
we need to construct a large program search space that contains the tensorization.
We generalize existing hierarchical search space generation~\cite{zheng2020ansor} to tensor computations. We construct the search space by generating program sketches that contain the tensorized computation, then enumerate over choices induced by 
the generated sketches. As shown in the right part in \autoref{fig:auto-tensorize}, a program sketch fixes parts of program structures while leaving space for remaining choices of parameters such as loop tiling size and computation caching decisions.
We generate sketches by applying pre-defined sketch generation rules iteratively. 
Importantly, we need to build
sketch generation rules that work on tensorized computations by looking at the block signatures and 
make use of the access region information during our analysis.

\paragraph{Data Movement as First-Class Citizen}  Existing auto-schedulers for tensor programs focus their designs on the schedule of computations and treat data movement between different memory scopes with secondary priority. However, since tensor intrinsics vastly improve the throughput of computations, data movements become the bottleneck of tensor programs.
Moreover, data movement decisions usually depend on computation schedule decisions like tilings, thread bindings, execution scopes, and producer-consumer data flow granularity. 
We take these insights and bring data movements as first-class citizens in our automatic scheduler and decouple them from computation schedules.
Specifically, we insert \textit{AutoCopy} blocks into the places where the sketch generation rules decide to perform data movements (\autoref{fig:auto-tensorize}). The copy block hides the memory schedule details and only exposes the necessary buffer access information at the block signature level. The isolated copy blocks allow the sketch generation to independently make computation schedule decisions without considering how to do data movements. The body of the \textit{AutoCopy} block describes the details of the data movement task, including buffer position mapping, threading, and storage scope requirements. A data movement scheduler takes this information as input and performs memory-related schedule transformations, such as inserting intermediate cache stages, utilizing data movement tensor intrinsics, vectorization, cooperative fetching, or stride padding to avoid bank conflicts.

\subsection{Evolutionary Search}

After the tensorized program sketch generation phase, we can get billions of possible induced programs. We use evolutionary search to explore the space and find an optimized tensorized program.
Our search starts from random initializations of choices for given program sketches. We then perform mutations on the current set of programs. We then select promising programs from the mutated candidates and benchmark them on our hardware backend of interest. We collect data from the evaluation phase to update the learned cost model.

\paragraph{Cost Model for Tensorized Computation} 
We build a boosting tree ensemble~\cite{XGBoostKDD} based cost models that use features extracted from the program. 
The feature vector contains information related to memory access patterns,
reuse, and loop annotations. Importantly, we extract features from both
block signatures in an isolated way as well as the body of the block~(e.g., to mark the use of Tensor Core). Our cost model can be viewed as a generalization of previous approaches
to tensorized programs. We believe an effective cost model for tensorized programs is a promising area for future research. 

\paragraph{Validation}

Randomly mutating programs during the search can generate invalid programs due to the unmet constraints 
of tensor intrinsic or invalid loop nest candidates. 
The possibility of false positives necessitates a validation step during the search.
We apply techniques in \autoref{subsec:validation} to validate a program candidate in the evolutionary search to identify and reject invalid programs. The validation step reduces the burden on evolutionary search algorithms and allows us to generate a small number of false positives during the search.
\section{Evaluation} \label{sec:evaluation}

We implement \tir{} on top of Apache TVM~\cite{chen2018tvm}. Notably, the insights described in the paper can benefit other machine learning compilation frameworks as well. This section provides evaluations to answer the following questions:

\begin{itemize}
    \item Can \tir{} optimize common set of machine learning operators ~(\S\ref{subsec:eval_single_op})? 
    \item Can \tir{} bring performance boost to end-to-end network execution~(\S\ref{subsec:eval_e2e})?
    \item Can \tir{} support tensor intrinsics on different hardware platforms~(\S\ref{subsec:eval_arm})?
\end{itemize}

To evaluate \tir{} along those axes, we compare our solution to existing machine learning compilation solutions on GPU and CPU platforms. We will discuss the specific setups in the corresponding subsections.

Additionally, we are interested in the following question throughout all evaluations: How does \tir{} compare to vendor-specific libraries and frameworks that rely on these libraries?  Importantly, most of these libraries are heavily optimized by a dedicated team of engineers. In all of these settings, \tir{} performs end-to-end automatic optimization without the need to call into external libraries. 


\subsection{Single Operator Evaluation}
\label{subsec:eval_single_op}

This section evaluates \tir{} on operators in deep learning models.
We pick a common collection of workloads, including: 1D convolution (C1D), 2D convolution (C2D), 3D convolution (C3D), depth-wise convolution (DEP), dilated convolution (DIL), general matrix multiply (GMM), group convolution (GRP), and transposed 2D convolution (T2D). The evaluations are done on an NVIDIA RTX 3080 platform with Tensor Cores. We pick this platform as it has a wide spectrum of machine learning compiler solutions and libraries that we can use as comparison reference points. We use float16 as the input and accumulator data type for all operators. We include TVM (commit: 27b0aad5, with auto-scheduler~\cite{zheng2020ansor}) and AMOS (commit: 6aee6fe2) as two machine learning compiler baselines. Finally, we compare against two vendor-specific solutions: CUTLASS (version 2.9) and TensorRT~(PyTorch-TensorRT container Release 22.06).

\paragraph{Comparisons to Machine Learning Compilers.}

\begin{figure}[t]
    \centering
    \includegraphics[width=\linewidth]{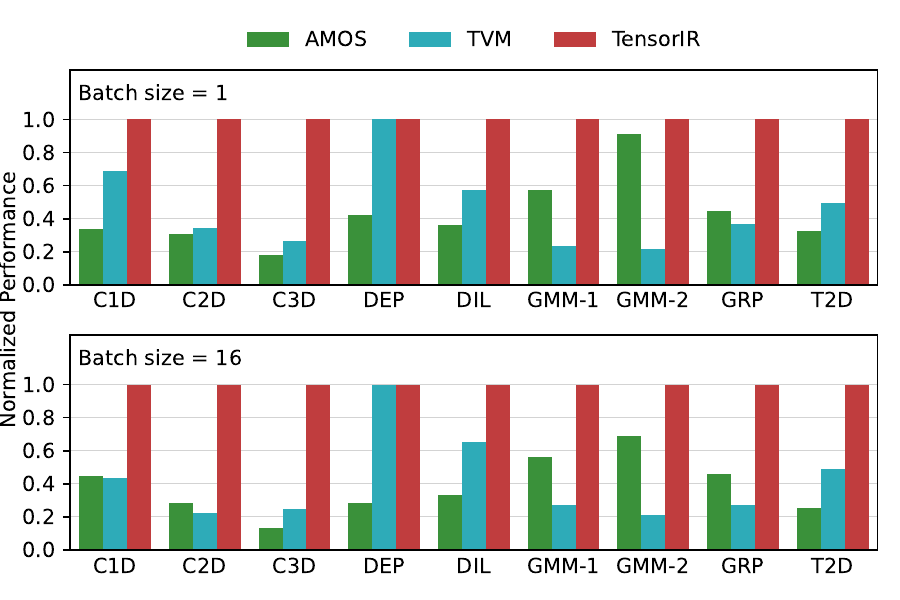}
    \precap
    \caption{Single operator comparison to existing machine learning compilers on Nvidia GPU. \tir{} brings up to 7.5x speed across workloads.}
    \postcap
    \label{fig:single_op_compiler}
\end{figure}

\autoref{fig:single_op_compiler} shows the comparisons to AMOS and TVM. 
TVM~\cite{chen2018tvm} works well on less compute-intensive workloads (\eg DEP), but has poor performance on heavy ones (\eg C2D, C3D, GMM) due to the limited Tensor Core support. AMOS~\cite{zheng2022amos} can use Tensor Core for every workload but not doing as well as \tir{}. Overall, \tir{} brings up to $7.5\times$ number of improvement over existing machine learning compilation solutions. These improvements come from better abstraction and automatic scheduling that leverages tensor compute intrinsics and corresponding data movements.

\paragraph{Comparisons to Platform Specific Libraries.}

\begin{figure}[t]
    \centering
    \includegraphics[width=\linewidth]{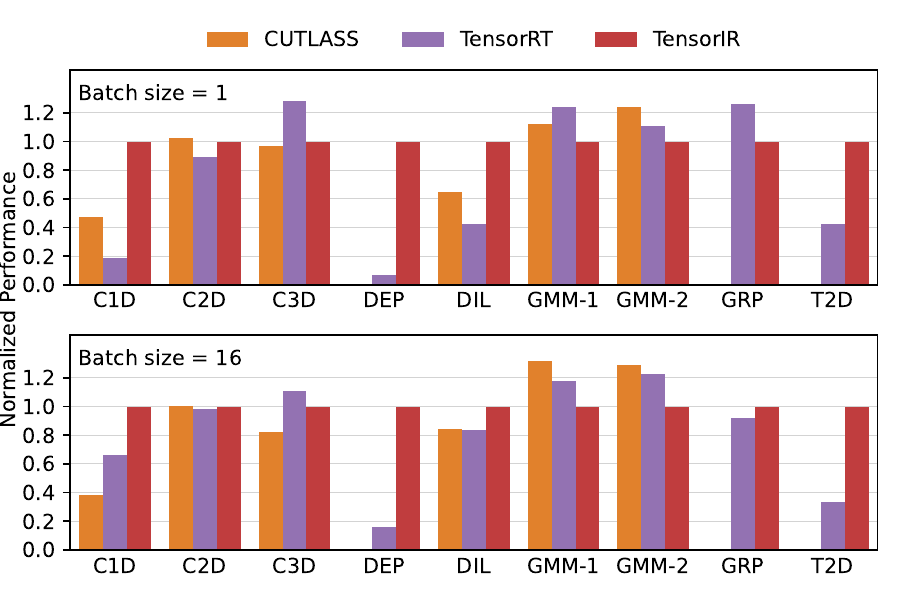}
    \precap
    \caption{Single operator comparison to platform-specific libraries.
    We did not show the numbers of CUTLASS on DEP, GRP, and T2D as the library does not support them.
    \tir{} outperforms the baselines on C1D, C2D, DEP, T2D, and DIL by up to 13.9x and gets to more than 75\% throughput on C3D, GMM, and GRP.}
    \postcap  
    \label{fig:single_op_library}
\end{figure}

\autoref{fig:single_op_library} shows the comparisons of \tir{} to two platform specific solutions:
CUTLASS~\cite{Kerr_CUTLASS_2022} and TensorRT~\cite{nvidia2017tensorrt}. 
\tir{} outperforms the baselines on C1D, C2D, DEP, T2D, and DIL by up to $13.9\times$. These results
shows the advantage of automatic optimizations provided by \tir{}.
Notably, \tir{} gets to more than $75\%$ on C3D, GRP an GMM. These results show that even on workloads that are intensively optimized by dedicated engineering teams, \tir{} can still get close to or match existing vendor-specific solutions. We expect the remaining gap continues to close as we bring additional insights from these libraries to \tir{}.
In all cases, the baseline solutions are optimized by dedicated engineering teams, while \tir{} enables automated compilation for a given tensor intrinsic declaration.




\subsection{End-to-end Model Evaluation}
\label{subsec:eval_e2e}


In this section, we evaluate the impacts that \tir{} can bring to end-to-end model execution.
We evaluate our solutions on four widely-used models \cite{he2016deep, howard2017mobilenets, devlin2018bert, dosovitskiy2020image} on on NVIDIA RTX 3080. We include 
 TVM  and AMOS as machine learning compiler baselines. Additionally, we also 
 include PyTorch (version 1.13.0.dev20220612) as a framework reference point. Finally, we include TensorRT, which is a vendor-specific solution that is heavily optimized by engineering teams at Nvidia.
 
\begin{figure}[t]
    \centering
    \includegraphics[width=\linewidth]{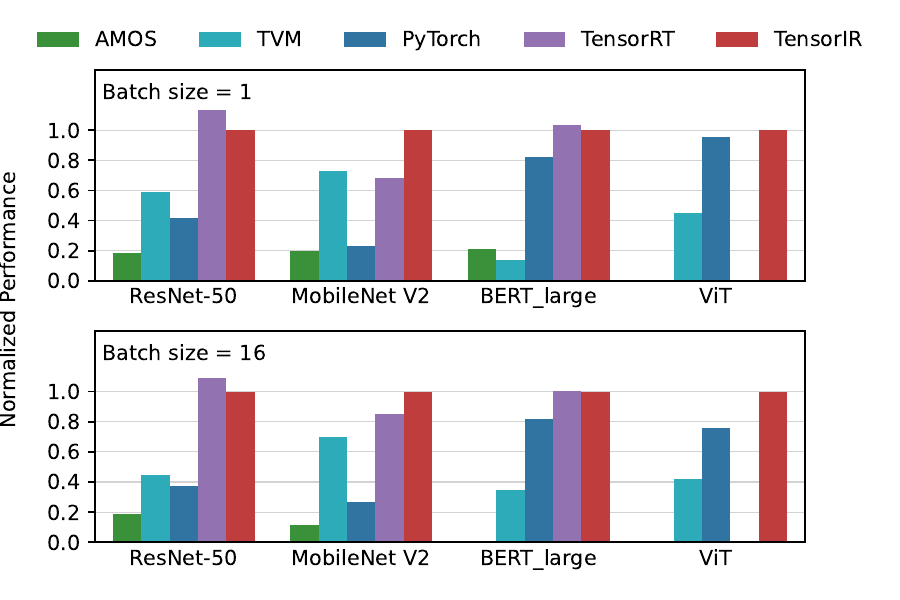}
    \precap
    \caption{End-to-End model evaluations on NVIDIA GPU. \tir{} significantly outperforms existing machine learning compilation solutions and achieves similar or better throughputs on popular networks compared with the inference libraries on GPUs. \tir{} get better performance on ViT, an emerging model that TensorRT does not yet support.}
    \postcap
    \label{fig:e2e_network}
\end{figure}

The results are shown in \autoref{fig:e2e_network}. \tir{} outperforms PyTorch, TVM, and AMOS by $1.2 - 8.8\times$. Additionally, \tir{} brings $30\%$ better performance on MobileNet V2 comparing to TensorRT, and achieves the $88\% - 100\%$ throughput on ResNet50 and BERT\_large.
Additionally, \tir{} can automatically support emerging models such as Vision Transformer, which TensorRT does not yet support. These results show that our abstraction and the automatic scheduler can bring close or even better performance than the best effort libraries on common machine learning models. Additionally, the automated solution enables us to bring faster support for emerging models.

Tuning time is an important factor of practical usability for search based automatic deep learning compilers. We compare the end-to-end tuning time of TVM and \tir{} which is shown in \autoref{table:time}. Our framework tunes up to 2x faster compared with TVM, and this improvement comes from two aspects. Firstly, hardware profiling contributes the most to the tuning time, and auto tensorization of \tir{} generates faster programs compared with TVM due to the utilization of Tensor Core and hence the profiling time is less accordingly. Secondly, our divide and conquer approach divides and isolates the problem space into outer loop nests and inner body. The inner body is tensorized with hardware intrinsics and we search over the loop transformations of outer loops, which results in a smaller search space. The search time can be tolerated when we deploy these models to many devices for months in production. \tir{} can eliminate search time further by caching historical cost models and search records. So no search is needed to build a model for an operator already tuned.


\subsection{ARM CPU Evaluation}
\label{subsec:eval_arm}

The last two subsections evaluate \tir{} on an Nvidia GPU. In this section, we study how easy it is to generalize \tir{} to different platforms. We evaluate results on an ARM platform by providing the description with 8-bit integer dot(\textit{sdot}). This instruction is different from the Tensor Core used in the last two subsections. Importantly, we use the same \tir{} framework by providing the new description of the tensor intrinsic to the system. The evaluations are done on an AWS Graviton2 CPU.

\begin{figure}[t]
    \centering
    \includegraphics[width=\linewidth]{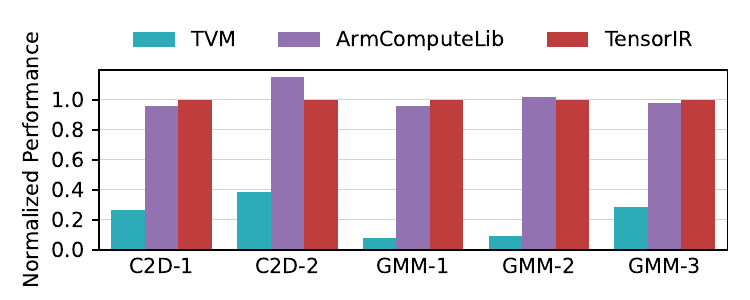}
    \precap
    \caption{Single operator evaluations on ARM CPU. \tir{} get up to 12.6x faster than TVM due to the use of native tensor instrinsic acceleration. It also gets to the same level of performance as heavily optimized platform specific library~(ArmComputeLib).}
    \postcap
    \label{fig:eval_arm_op}
\end{figure}
 
\begin{table}
\centering
\begin{tabular}{ |p{2cm}|p{2cm}|p{2cm}| }
 \hline
 \multicolumn{3}{|c|}{Tuning time} \\
 \hline
 Model& TVM (min) & \tir{} (min) \\
 \hline
 ResNet-50   & 308    & 156 \\
 MobileNet-v2  & 292   & 261 \\
 BERT          & 410  & 189 \\
 ViT           & 247  & 145 \\
 \hline
\end{tabular}
\caption{Tuning time comparison of end-to-end models on NVIDIA GPU. \tir{} tunes up to 2x faster.}
\label{table:time}
\end{table}
 
\paragraph{Single Operator Results.} We evaluate the results on two commonly used operators: C2D and GMM. We include TVM as a machine learning compiler baseline and ARMComputeLib~\cite{acl} as a platform-specific library baseline. The results are shown in \autoref{fig:eval_arm_op}.  \tir{} achieves up to $12.5\times$ speed up compared with TVM thanks to the ability to leverage native hardware acceleration. In the meantime, \tir{} reaches $85\% - 105\%$ throughput of ARMComputeLib~\cite{acl}, showing our ability to get to the same level of performance as vendor-specific solutions on this platform.

\begin{figure}[t]
    \centering
    \includegraphics[width=\linewidth]{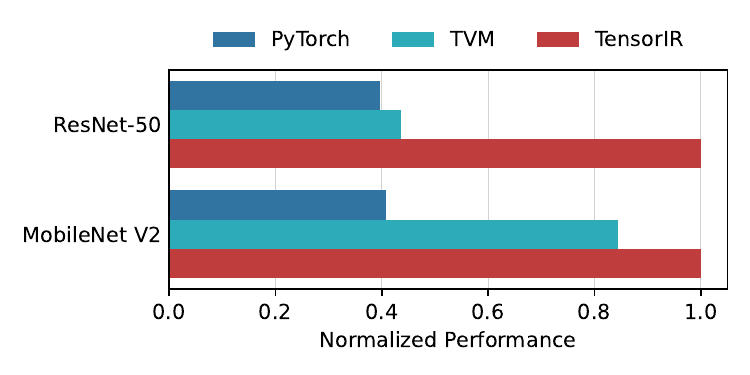}
    \precap
    \caption{End-to-end evaluation results on ARM CPU. \tir{} outperform with 1.2x--2.5x than PyTorch and TVM.}
    \postcap
    \label{fig:eval_arm_e2e}
\end{figure}

\paragraph{End-to-End Results.} Finally, we evaluate the end-to-end neural network executions on this platform. Our baselines include PyTorch and TVM. We achieve up to $2.3\times$ on this platform in \autoref{fig:eval_arm_e2e}. Notably, PyTorch contains a specialized quantized model support with QNNPACK~\cite{qnnpack} backend. However, QNNPACK has not yet added \textit{sdot} support. This result alludes to the maintenance cost for these frameworks to keep up with hardware changes. \tir{} can help to reduce the maintenance burden through automation and still bring competitive performance to hand-optimized systems.
\section{Related Works}
\label{sec:related-works}
Deep learning frameworks~\cite{abadi2016tensorflow,paszke2019pytorch, chen2015mxnet} optimize deep neural networks by invoking vendor optimized libraries (\eg, cuDNN \cite{chetlur2014cudnn}, MKL-DNN \cite{intel2017mkldnn}, TensorRT \cite{nvidia2017tensorrt}, ArmComputeLibrary~\cite{acl}). Libraries have engineering development costs and are specific to a particular hardware. \tir{} complements library developments to enable better coverage and reduce development costs by automatically providing comparable performance with vendor libraries. These frameworks
can leverage \tir{} to generate optimized tensorized programs for various hardware 
backends.

Compute-intensive linear algebra operators such as matrix multiplication and dot products has been a long-standing optimization target in HPC community (e.g., CUTLASS~\cite{Kerr_CUTLASS_2022}) due to their importance in scientific computation. The divide-and-conquer is a typical optimization technique in HPC and ML engineering. \tir{} takes these ideas and generalizes them to an abstraction that allows automatic tensorization.

Machine learning and tensor compilers introduce different abstractions for tensor programs.
Halide~\cite{ragan2013halide} and TVM~\cite{chen2018tvm} use a scheduling 
language that can describe loop optimization primitives of loop nests with a scalar body.
Tensor Comprehensions~\cite{vasilache2018tensor}, Tiramisu~\cite{baghdadi19tiramisu} 
and MLIR/Affine~\cite{mlir} use polyhedral model~\cite{vasilache2006polyhedral} to analyze loop nest dependencies. These works optimize loop nests with scalar computation 
in a bottom-up way. 
Fireiron~\cite{hagedorn2020fireiron} and Stripe~\cite{zerrell2019stripe} use
nested polyhedral structures to model tensor programs in a top-down fashion. 
\tir{} combines insights from both approaches and generalizes the representation
to tensorized programs. 
IREE~\cite{iree} is a compiler chain for end-to-end compilation flow which utilizes platform-specific optimization pipelines. \tir{} focuses on automating the tensorization process to generate optimized code for multiple platforms without human intervention.
TACO~\cite{kjolstad:2017:taco, chou:2020:conversion,senanayake:2020:scheduling}
is a compiler for sparse tensor algebra. Cortex~\cite{CortexMLSys21}  generalized
tensor compilation to recursive computations. Our work is orthogonal to these efforts.
We believe the \tir{} abstraction can be combined with insights from these works in the future to enable an even broader range of computations.

Automation is an essential topic in machine learning compilation and tensor program optimization. AutoTVM~\cite{chen2018learning} introduced a learning-based approach
to optimize tensor programs via a learned cost model and template-guided search. Triton~\cite{Triton} introduces
a tile-based template representation for effective program optimization.
FlexTensor~\cite{zheng2020flextensor} automatically generates the template.
Halide builds an automatic scheduler using Monte-Carlo tree search~\cite{adams2019learning}. 
Ansor~\cite{zheng2020ansor} improves automatic scheduling using a hierarchical 
search space. Our automatic scheduling algorithm takes lessons from these approaches and
generalizes them to tensorized computation best for domain-specific hardware acceleration.

Auto-vectorization~\cite{rosen2007loop, kong2013polyhedral} is a long-standing topic in compiler research. Tensorization can be viewed as a generalization of the vectorization
problem to enable tensor intrinsic in modern accelerators~\cite{nvidia2017tensorcore, intel2019vnni, arm2017dot, moreau2018vta}. There are some existing works\cite{bhaskaracharya2020automatic, zhao2021akg, weng2021unit, zheng2022amos} on this topic. AKG~\cite{zhao2021akg} uses the polyhedral method to explore tensorized search space, UNIT~\cite{weng2021unit} introduces a generic flow for tensorization, 
while AMOS~\cite{zheng2022amos} enables automatic mapping to tensorized intrinsic through tensor expression. Our method generalizes these previous approaches by proposing a novel abstraction for tensorization computation and jointly performing tensorization along with other optimizations. \tir{} serves as a foundation
to further develop tensorization-aware automatic scheduling methods.
\section{Conclusion}

We propose \tir, an abstraction for automatic tensorized program optimization. We design a 
key abstraction called block that can isolate tensorized computations
and provide effective transformation primitives for program optimization.
We build an automatic scheduling algorithm that performs tensorization jointly with
other optimizations and generates performant programs.
We hope this work will encourage additional studies of
tensorized program optimization and provide new opportunities for
hardware and software specialization.

\begin{acks}
This work is supported in part by a gift from Oppo. We would like to thank Masahiro Masuda from OctoML and the TVM community for helpful discussion and support of the work. We would like to thank anonymous reviewers and our shepherd Christian DeLozier for their helpful feedback and discussions.
\end{acks}

\bibliographystyle{ACM-Reference-Format}
\bibliography{references}

\end{document}